%% file: neurips_2026.tex
\documentclass{article}

\usepackage[preprint]{neurips_2026}

\usepackage[utf8]{inputenc} 
\usepackage[T1]{fontenc}    
\usepackage{hyperref}       
\usepackage{url}            
\usepackage{booktabs}       
\usepackage{amsfonts}       
\usepackage{nicefrac}       
\usepackage{microtype}      
\usepackage{xcolor}         
\usepackage{amsmath}
\usepackage{multirow} 
\usepackage{booktabs}  

\usepackage{graphicx} 
\title{Memento: Reconstruct to Remember for Consistent Long Video Generation}

\author{%
\textbf{Xuan Wei}$^{1, 2,\ddagger}$ \quad
\textbf{Longbin Ji}$^{2}$ \quad
\textbf{Guan Wang}$^{2}$  \quad
\textbf{Xiangrui Liu}$^{2,\ddagger}$  \quad
\\
\textbf{Zhenyu Zhang}$^{2,\S}$ \quad
\textbf{Shuohuan Wang}$^{2}$ \quad
\textbf{Yu Sun}$^{2}$ \quad
\textbf{Qingqi Hong}$^{1,\dagger}$ \quad
\\
[2mm]
$^{1}$Xiamen University \quad $^{2}$ERNIE Team, Baidu Inc. \\
[1mm]
\texttt{weixuan@stu.xmu.edu.cn}, \texttt{\{jilongbin,zhangzhenyu07\}@baidu.com} \\
[1mm]
{\small $^{\ddagger}$Work done during internship at Baidu. \quad 
$^{\S}$Project Lead. \quad 
$^{\dagger}$Corresponding author.} \\
}

\begin{document}

\maketitle

\begin{abstract}
Long-form video generation requires recurring subjects to remain consistent across various shots, viewpoints, motions, and scene transitions.
Existing temporal decomposition methods improve scalability by generating videos shot by shot.
However, they mainly focus on optimizing plausible next-shot continuations without verifying whether the historical memory preserves identity-critical subject evidence. 
Consequently, as generation proceeds, recurring subjects may be diluted, overwritten, or forgotten.
In this paper, we propose \textbf{Memento}, a subject-reconstruction-guided framework that treats subject preservation as an explicit identity grounding problem, based on the premise that a memory bank faithfully preserving a subject should support reconstructing that subject from memory alone.
Specifically, Memento jointly trains autoregressive next-shot generation with memory-based subject reconstruction, recovering target appearances using historical memory and global story captions.
To disentangle long-range subject evidence from short-range cues, Memento introduces a dual-query memory mechanism, where one query retrieves identity-relevant memory and the other selects short-context keyframes for coherent continuation. 
Additionally, a subject-aware cinematic data pipeline provides precise reconstruction supervision via consistent, pronoun-free subject descriptions.
Experiments demonstrate that Memento achieves state-of-the-art performance in long-term subject consistency, cross-shot coherence, and visual quality.
\end{abstract}

\input{section/intro}

\section{Related Work}
\label{gen_inst}

\subsection{Single-Shot Video Generation}

Recent advances in diffusion models~\cite{diffusion} and Diffusion Transformers
(DiTs)~\cite{dit} have substantially improved video generation. Large-scale
text-to-video (T2V) and image-to-video (I2V) models~\cite{singer2022make,
ho2022video, blattmann2023stable, wan, kong2024hunyuanvideo,
yang2024cogvideox, veo3, kling, sora2, yu2025context, seedance2026seedance}
can now synthesize short, single-shot clips with high visual fidelity and natural
motion. However, these models are primarily designed for temporally bounded
single scenes and lack explicit mechanisms for maintaining consistency across
multiple shots. Consequently, they remain insufficient for long-form,
story-driven video generation, where recurring subjects, scenes, and narrative
logic must remain coherent over extended sequences. This limitation motivates
the development of multi-shot video generation approaches.

\subsection{Multi-Shot Long Video Generation}
\label{sec:related_multi_shot}

Multi-shot video generation aims to produce long-form narratives by decomposing video synthesis into shot-level generation. Existing methods mainly follow three paradigms.

\noindent\textbf{Storyboard-based generation.}
Storyboard-based methods~\cite{xie2024dreamfactory, zhao2024moviedreamer,
zheng2024videogen, hu2024storyagent, storydiffusion, he2025cut2next,
xiao2025captain, zhang2025shouldershot, long2024videostudio, iclora}
first generate sparse keyframes and then animate them with pretrained image-to-video models. Although efficient, they enforce consistency mostly at the keyframe level, leaving individual clips weakly coupled and often causing rigid transitions, detail drift, and limited narrative coherence.

\noindent\textbf{Joint multi-shot generation.}
Joint generation methods synthesize multiple shots within a single diffusion forward pass. Representative works such as LCT~\cite{lct} and HoloCine~\cite{holocine} introduce cross-shot attention mechanisms to improve global interaction and shot-level alignment, with later methods further improving efficiency and controllability~\cite{wu2025cinetrans, qi2025mask, kara2025shotadapter, cai2025mixture, jia2025moga}. However, these methods remain bounded by the context window, and their memory and computation costs increase rapidly as the number of shots grows, limiting open-ended generation.

\noindent\textbf{Memory-conditioned autoregressive generation.}
Autoregressive methods scale better by generating videos shot by shot while conditioning on a compact memory of previous content. StoryMem~\cite{storymem} maintains selected keyframes using semantic and aesthetic criteria, while OneStory~\cite{onestory} compresses historical context for next-shot generation. These designs improve scalability, but their memory selection is primarily based on generic relevance or short-term context compression, rather than explicit subject preservation. As a result, identity-critical cues may be diluted or overwritten over long generation horizons.

In contrast, \textbf{Memento} explicitly formulates long-form consistency as subject grounding in memory. It uses a dual-query memory mechanism to separate long-context subject evidence from short-context generation cues, and jointly optimizes next-shot generation with memory-based subject reconstruction. This encourages the memory to retain identity-critical information while preserving local visual continuity, enabling more stable long video generation.

\section{Method}
\label{headings}

\begin{figure*}[t]
    \centering
    \includegraphics[width=\textwidth]{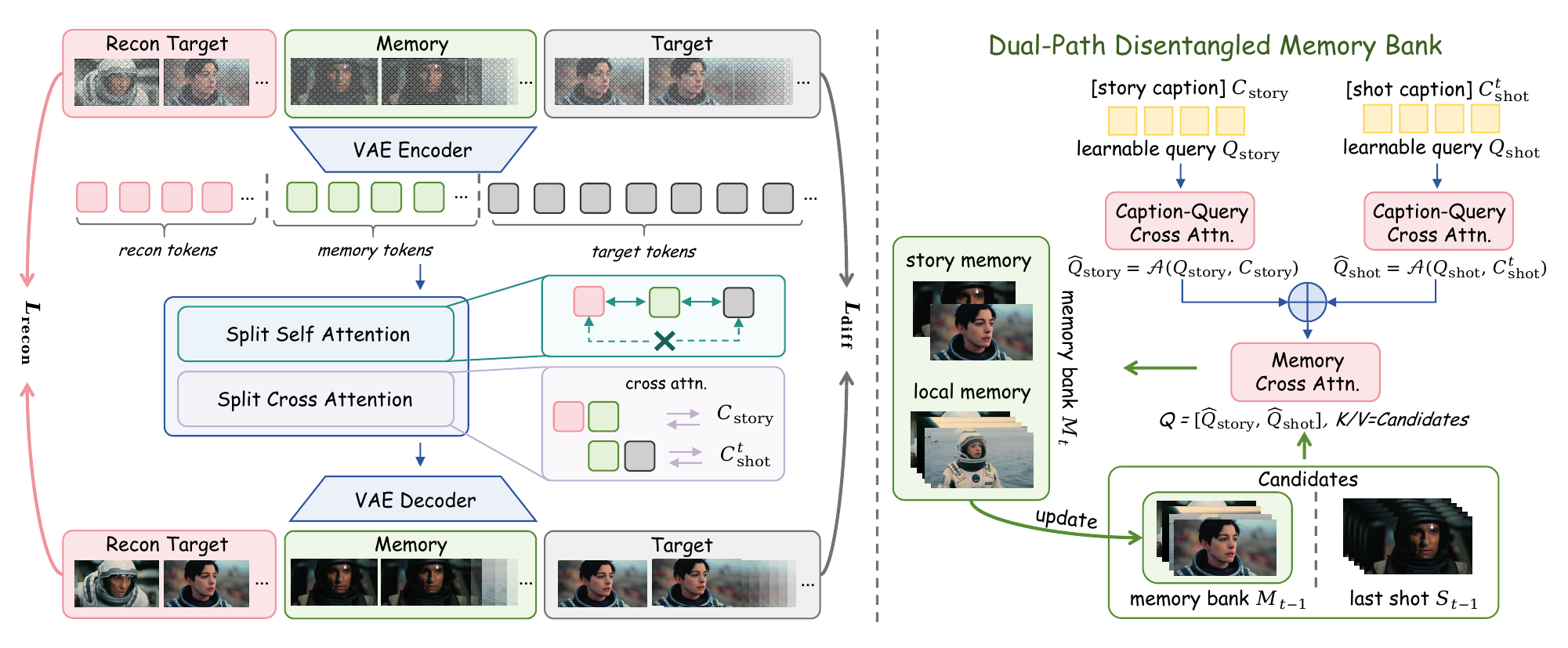} 
    \caption{
    Overview of our framework. The left side illustrates the generation process, while the right side details the memory bank update mechanism. We employ split self-attention over overlapping local groups, allowing reconstruction-memory and memory-target interactions while avoiding full global attention. Split cross-attention injects both global story-level and local shot-level captions. Story and shot captions condition separate learnable queries via caption-to-query cross-attention. The fused query retrieves relevant candidates from the memory bank and the last shot, and updates global and local memory states for scalable long-form generation.
}
    \label{fig:qualitative_compare}
\end{figure*}

Memento follows the principle that consistent long video generation requires both persistent subject evidence and recent contextual cues. To this end, we generate videos autoregressively at the shot level with a compact historical memory bank, and explicitly supervise the memory through subject reconstruction. This design allows the model to preserve recurring identities without attending to all previous frames, while still using recent visual context for local shot continuity.

Given a global story caption $C_{\mathrm{story}}$ and a sequence of per-shot captions $\{C_{\mathrm{shot}}^t\}_{t=1}^{N}$, Memento generates a multi-shot video $V={S_1,\dots,S_N}$. The first shot is generated by standard text-to-video synthesis. For each subsequent shot $t \geq 2$, the model performs memory-conditioned generation using $C_{\mathrm{story}}$, $C_{\mathrm{shot}}^t$, and a dynamically updated memory bank $M_t$. Optionally, the last frame of $S_{t-1}$ can be provided for smoother inter-shot transitions:
\begin{equation}
    S_t = \theta\!\left(M_{t},\, C_{\mathrm{story}},\, C_{\mathrm{shot}}^t, S_{t-1}\right)
\end{equation}
where $S_{t-1}$ is omitted in the Text-and-Memory-to-Video (TM2V) generation setting. The framework consists of three components: a subject-aware data curation pipeline that provides unambiguous subject supervision, a dual-query memory mechanism that retrieves long-context subject evidence and short-context generation cues, and a subject-anchored multi-task objective that jointly trains next-shot generation with memory-based subject reconstruction.

\subsection{Subject-Aware Data Curation Pipeline}
\label{sec:data}

\begin{figure}[t]
\centering
\includegraphics[width=0.9\textwidth]{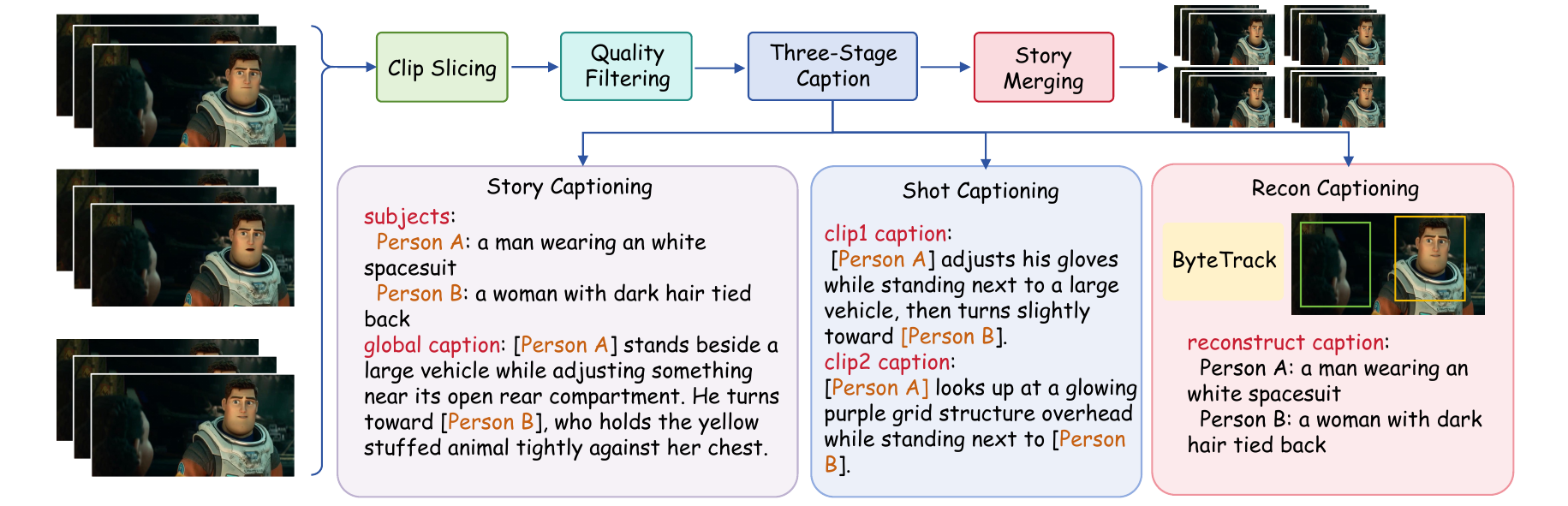}
\caption{Overview of the subject-aware data curation pipeline. We design three-stage captions to support consistent long video generation. Story Caption generates a global narrative and identifies all recurring subjects with fixed, pronoun-free names. Shot Caption produces per-shot descriptions that adhere to the established subject inventory. Recon Caption annotates selected subject-salient frames as reconstruction targets for memory-based identity supervision.}
\label{fig:data}
\end{figure}

Training Memento requires annotations that support both shot-level generation and memory-based subject reconstruction. Beyond conventional video-caption pairs, the data must consistently identify recurring subjects across shots and provide reconstruction targets that contain clear visual evidence of their appearances. However, standard captions often describe subjects with ambiguous references such as `the man,'' `he,'' or ``she,'' making it difficult to associate the same subject across changing scenes, viewpoints, and interactions. To address this issue, we construct a subject-aware data curation pipeline that produces temporally coherent multi-shot videos with explicit, pronoun-free subject descriptions and reconstruction targets.

As shown in Fig.~\ref{fig:data}, we collect raw footage from complete cinematic productions, including movies, documentaries, and animations. We slice the footage into continuous multi-shot sequences of 30--60 seconds, where each sequence contains multiple natural shots and each shot lasts longer than 5 seconds. Before captioning, we apply quality filtering and crop subtitle regions to reduce visual artifacts that may degrade generation quality.

\noindent\textbf{Stage 1: Story Captioning.}
We first process the full multi-shot sequence with Qwen3-VL~\cite{qwen3} to generate a global story caption and identify salient recurring subjects throughout the sequence. This stage establishes a subject inventory shared by all subsequent annotations, ensuring that each recurring subject is described with a fixed and explicit textual reference. In parallel, we apply ByteTrack~\cite{zhang2022bytetrack} to track the primary subject over time and select two frames where the subject occupies the largest visual area as reconstruction targets. These frames provide high-quality appearance evidence for the reconstruction branch.

\noindent\textbf{Stage 2: Shot Captioning.}
Given the global story caption and the subject inventory from Stage~1, we caption each individual shot using Qwen3-VL. Instead of allowing generic noun phrases or pronouns, we enforce the fixed subject descriptions when referring to recurring subjects. This aligns local shot captions with the global subject identities and reduces ambiguity in cross-shot subject association, especially when multiple visually similar subjects appear in the same sequence.

\noindent\textbf{Stage 3: Reconstruction Target Captioning.}
Finally, we caption the selected reconstruction target frames using Qwen3-VL. The generated captions are constrained to remain consistent with the global story caption and the established subject descriptions, so that each target image is paired with an unambiguous subject reference. These target image-caption pairs are used to supervise memory-based subject reconstruction, encouraging the memory bank to preserve identity-relevant appearance cues rather than only short-term visual context.

\subsection{Dual-Path Disentangled Memory Bank}
\label{sec:memory}

A fixed-size memory is necessary for scalable autoregressive long video generation, but a single memory selection criterion can entangle different types of historical information. In our setting, subject reconstruction requires long-range identity evidence, while next-shot generation relies more on recent scene context. We therefore design a dual-path memory mechanism that retrieves subject-relevant and shot-relevant memories separately from a shared candidate pool.

\noindent\textbf{Memory bank construction.}
At generation step $t$, Memento maintains a fixed-length memory bank $M_t$ that summarizes useful historical visual information. Before generating the $t$-th shot, we form a candidate memory pool by combining the previous memory bank $M_{t-1}$ with visual features extracted from the most recently generated shot $S_{t-1}$:
\begin{equation}
M_t^{\mathrm{cand}} =
\left[M_{t-1}\;;\;\mathcal{E}(S_{t-1})\right]
\in \mathbb{R}^{N_{\mathrm{cand}} \times D}
\end{equation}
where $\mathcal{E}(\cdot)$ denotes a pretrained VAE encoder that maps frames into latent features of dimension $D$, and $N_{\mathrm{cand}}$ is the number of candidate memory tokens. This construction allows the memory to retain selected information from earlier shots while continuously incorporating newly generated visual evidence.

\noindent\textbf{Dual-query disentanglement.}
To separate identity-oriented retrieval from context-oriented retrieval, we introduce two sets of learnable query tokens: story queries $Q_{\mathrm{story}} \in \mathbb{R}^{N_{\mathrm{story}} \times D}$ and shot queries $Q_{\mathrm{shot}} \in \mathbb{R}^{N_{\mathrm{shot}} \times D}$. The story queries are conditioned on the global story caption to retrieve subject evidence that should remain stable across shots, while the shot queries are conditioned on the current shot caption to retrieve local references useful for the upcoming generation:
\begin{equation}
\hat{Q}_{\mathrm{story}} =
\mathcal{A}\!\left(Q_{\mathrm{story}},\, \phi(C_{\mathrm{story}})\right),
\qquad
\hat{Q}_{\mathrm{shot}} =
\mathcal{A}\!\left(Q_{\mathrm{shot}},\, \phi(C^t_{\mathrm{shot}})\right)
\end{equation}
where $\phi(\cdot)$ encodes text conditions into the shared latent space and $\mathcal{A}(\cdot,\cdot)$ denotes cross-attention. Through this conditioning, $\hat{Q}_{\mathrm{story}}$ is guided by the global narrative and recurring subject descriptions, whereas $\hat{Q}_{\mathrm{shot}}$ focuses on the local semantics of the current shot.

\noindent\textbf{Adaptive memory selection.}
The two query sets independently score the candidate memory pool:
\begin{equation}
s_{\mathrm{story}} =
\mathcal{S}\!\left(\hat{Q}_{\mathrm{story}},\, M_t^{\mathrm{cand}}\right),
\qquad
s_{\mathrm{shot}} =
\mathcal{S}\!\left(\hat{Q}_{\mathrm{shot}},\, M_t^{\mathrm{cand}}\right)
\end{equation}
where $\mathcal{S}(\cdot,\cdot)$ denotes the relevance scoring function. We then select the top-$K$ memory tokens for each path and concatenate the selected subsets to form the updated memory bank:
\begin{equation}
M_t =
\left[
\operatorname{TopK}\!\left(M_t^{\mathrm{cand}},\, s_{\mathrm{story}}\right)
\;;\;
\operatorname{TopK}\!\left(M_t^{\mathrm{cand}},\, s_{\mathrm{shot}}\right)
\right]
\end{equation}
where $\operatorname{TopK}(M, s)$ retrieves the top-$K$ tokens from $M$ according to score $s$. The story-selected memories provide long-context subject evidence for identity grounding and reconstruction, while the shot-selected memories provide short-context visual cues for local generation. By decoupling the two retrieval paths, the memory bank reduces competition between persistent identity information and transient scene context.

\subsection{Subject-Anchored Multi-Task Training}
\label{sec:training}

The dual-path memory bank provides candidate identity and context evidence, but next-shot generation alone does not guarantee that the selected memory contains sufficient subject information. We therefore introduce subject-anchored multi-task training, which couples standard next-shot video denoising with memory-based subject reconstruction. This auxiliary objective shares the memory-conditioned generator with next-shot denoising, adding identity supervision without extra training stages or inference-time overhead.

\noindent\textbf{Next-shot generation.}
The primary task is conditional diffusion denoising for the current shot. Given the noisy shot latent $S_t^\tau$ at diffusion timestep $\tau$, the model predicts the added Gaussian noise conditioned on the memory bank $M_t$, the global story caption $C_{\mathrm{story}}$, and the current shot caption $C_{\mathrm{shot}}^t$:
\begin{equation}
    \mathcal{L}_{\mathrm{diff}} =
\mathbb{E}\!\left[\,
\bigl\|\epsilon - \epsilon_\theta\bigl(S_t^{\tau},\tau,M_t,C_{\mathrm{story}},C_{\mathrm{shot}}^t\bigr)\bigr\|^2
\,\right]
\end{equation}
where $\epsilon \sim \mathcal{N}(0,I)$ denotes the added Gaussian noise. This objective trains the model to generate the next shot from both textual instructions and historical visual memory.

\noindent\textbf{Subject-role reconstruction.}
To make identity preservation directly optimizable, we formulate an auxiliary Text-and-Memory-to-Image (TM2I) reconstruction task. Given a target subject image $I_{\mathrm{sub}}$, the model predicts its diffusion noise conditioned only on the memory bank $M_t$ and the global story caption $C_{\mathrm{story}}$, without using any direct visual prompt:
\begin{equation}
    \mathcal{L}_{\mathrm{recon}} =
\mathbb{E}\!\left[\,
\bigl\|\epsilon - \epsilon_\theta\bigl(I_{\mathrm{sub}}^{\tau},\tau,M_t,C_{\mathrm{story}}\bigr)\bigr\|^2
\,\right]
\end{equation}
Since the reconstruction branch cannot access the target image except through its noisy latent, successful denoising requires the model to retrieve identity-relevant appearance cues from memory under the guidance of the story caption. This discourages the memory from serving only short-term generation context and anchors it to recurring subject identities.

\noindent\textbf{Total loss.}
The final training objective combines next-shot generation and subject reconstruction:
\begin{equation}
\label{eq_loss}
    \mathcal{L}_{\mathrm{total}} = \mathcal{L}_{\mathrm{diff}} + \lambda\,\mathcal{L}_{\mathrm{recon}}
\end{equation}
where  $\lambda$ is a balancing hyperparameter that controls the trade-off between scene progression and subject reconstruction fidelity.


\section{Experiments}

We conduct extensive experiments to evaluate the effectiveness of Memento in generating long videos with high subject consistency and temporal coherence. First, we provide implementation details on model training and inference (Sec.~\ref{Implementation}). Next, we provide a comprehensive evaluation against state-of-the-art multi-shot video generation methods through quantitative metrics, qualitative visual results, and a user study (Sec.\ref{Comparison}). Then, we investigate the individual contributions of our core components in Sec.\ref{ablation}. Finally, we show advanced capabilities of our method in Sec.\ref{capability}.

\subsection{Implementation Details}
\label{Implementation}

\textbf{Training Setup.} Our video generation backbone is built upon Wan2.2~\cite{wan} 14B. The training data is processed through our subject-aware data curation pipeline, resulting in 20K clips at a resolution of $480 \times 832$. We employ the AdamW~\cite{adamw} optimizer with a base learning rate of $1e-5$. The model is trained on $16$ NVIDIA H100 GPUs. The loss weight parameter $\lambda$ in Eq.~\ref{eq_loss} is empirically set to $0.5$. To facilitate robust classifier-free guidance (CFG) during inference, we randomly drop the textual conditions $C_{\mathrm{story}}$ and $C_{\mathrm{shot}}$ with a probability of $10\%$ during training.

\textbf{Inference Configurations.}
We follow the STBench autoregressive evaluation protocol and adopt the scene transition strategy of StoryMem~\cite{storymem}. The first shot $S_1$ is generated from Gaussian noise with standard T2V. For later shots, we use TMI2V within the same scene, injecting the last frame of the previous shot for pixel-level continuity, and switch to TM2V across scene changes, relying on text prompts and the updated memory bank $M_t$ for long-term identity preservation. We use the UniPC~\cite{zhao2023unipc} sampler with 40 denoising steps for both generation and reconstruction, and set the CFG scale for global and local text conditions to 3.5.


\subsection{Comparison}
\label{Comparison}
\subsubsection{Quantitative Results}

\begin{table*}[t]
    \centering
    \caption{Quantitative comparison of video generation capabilities. We report aesthetic quality, semantic consistency at both story and shot levels, background consistency, and subject consistency across different granularities. The best results are highlighted in \textbf{bold}, and the second-best results are \underline{underlined}.}
    \label{tab:comparison}



    \resizebox{\textwidth}{!}{
    \begin{tabular}{l c cc c ccc}
        \toprule
        \multirow{2}{*}{Method} 
        & \multirow{2}{*}{Aesthetic} 
        & \multicolumn{2}{c}{Semantic Consistency} 
        & \multirow{2}{*}{\shortstack{Background \\ Consistency}} 
        & \multicolumn{3}{c}{Subject Consistency} \\
        \cmidrule(lr){3-4} \cmidrule(lr){6-8}
        & & Story & Shot & & Inter-shot & Intra-shot & Inter-scene \\
        \midrule
        StoryDiffusion + Wan2.2-I2V~\cite{storydiffusion} 
        & \textbf{0.5310} & 0.2671 & 0.2689 & 0.9767 & 0.5525 & \underline{0.8448} & \underline{0.6732} \\
        StoryMem~\cite{storymem} 
        & 0.4937 & \underline{0.2793} & 0.2681 & 0.9732 & \underline{0.6606} & 0.8146 & 0.6692 \\
        HoloCine~\cite{holocine} 
        & 0.4568 & 0.2720 & \underline{0.2854} & \underline{0.9770} & 0.5791 & 0.8128 & 0.6594 \\
        \midrule
        Ours     
        & \underline{0.4977} & \textbf{0.3063} & \textbf{0.2893} & \textbf{0.9805}
        & \textbf{0.7338} & \textbf{0.8578} & \textbf{0.7268} \\
        \bottomrule
    \end{tabular}
    }
\end{table*}

We evaluate generated videos along four dimensions: aesthetic quality, semantic consistency, background consistency, and subject consistency. Subject consistency is measured at three temporal granularities: intra-shot, inter-shot, and inter-scene, capturing stability within a shot, across consecutive shots, and across scene transitions.

We compare Memento with representative methods from the three main paradigms: StoryDiffusion+Wan2.2-I2V~\cite{storydiffusion, wan} for storyboard-based generation, StoryMem~\cite{storymem} for memory-conditioned autoregressive generation, and HoloCine~\cite{holocine} for joint multi-shot generation. For HoloCine, we compute shot- and scene-level metrics at their corresponding granularities to avoid biases from temporal slicing.

As shown in Table~\ref{tab:comparison}, Memento achieves the best performance on long-term subject consistency. It obtains the highest inter-shot consistency of $0.7338$, outperforming StoryMem ($0.6606$), the highest intra-shot consistency of $0.8578$, surpassing StoryDiffusion+Wan2.2-I2V ($0.8448$), and the highest inter-scene consistency of $0.7268$, surpassing StoryDiffusion+Wan2.2-I2V ($0.6732$). These results indicate stronger identity preservation across both consecutive shots and scene transitions. Memento also achieves the best story-level semantic consistency ($0.3063$) and shot-level semantic consistency ($0.2893$), suggesting improved narrative coherence. Meanwhile, it obtains the second-best aesthetic quality ($0.4977$), showing that stronger subject preservation does not compromise visual fidelity or scene stability.

\subsubsection{Qualitative Results}

\begin{figure*}[t]
    \centering
    \includegraphics[width=\textwidth]{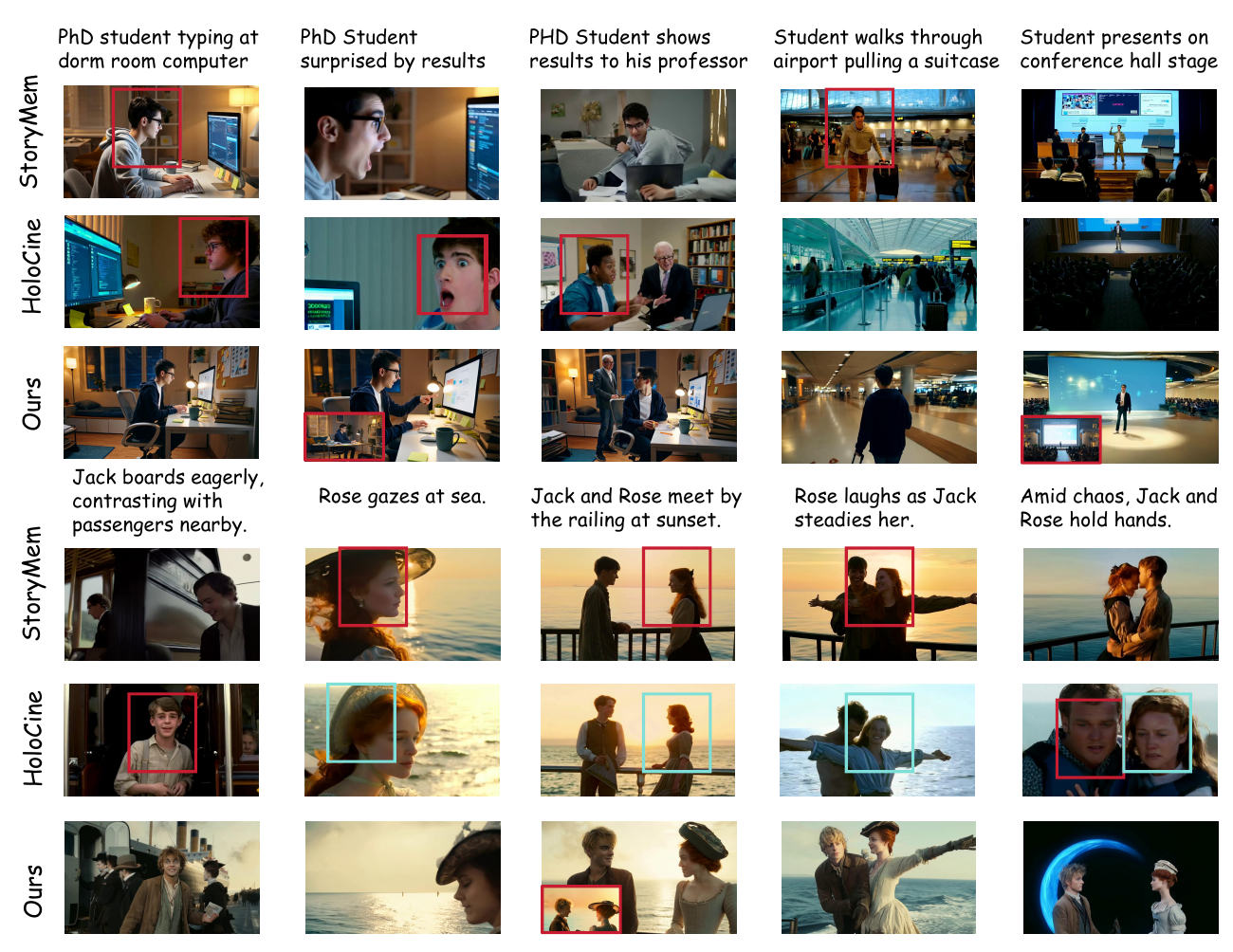} 
    \caption{Qualitative comparison of multi-shot video generation against StoryMem~\cite{storymem} and HoloCine~\cite{holocine}. Each row shows one method and each column shows a key story moment. Blue/red boxes highlight cross-shot subject inconsistencies, while red boxes also mark reconstruction frames. Our method better preserves subject identity across diverse scenes, viewpoints, and story progression.}
    \label{fig:qualitative_compare}
    \vspace{-0.2cm}
\end{figure*}

Fig.~\ref{fig:qualitative_compare} compares long multi-shot generation results from StoryMem~\cite{storymem}, HoloCine~\cite{holocine}, and our method. Each column shows a key story moment, and each row corresponds to one method. The highlighted regions indicate visible subject inconsistencies across shots.

StoryMem preserves identity in early shots but gradually drifts as generation proceeds, suggesting that its selected memory frames do not always retain subject cues required for future shots. HoloCine achieves reasonable within-scene consistency, but suffers from large identity changes across scene transitions, especially when viewpoint, background, or narrative context changes.

In contrast, our method maintains more stable subject appearance across both intra-scene motion and cross-scene transitions. The reconstruction frames highlighted in red provide explicit subject anchors, helping preserve identity under changes in camera view, lighting, background, and story context. These results show that subject-aware captioning and reconstruction-guided training effectively reduce identity drift and improve long-term narrative coherence.

\subsubsection{User Study}

\begin{figure*}[t]
    \centering
    \includegraphics[width=\textwidth]{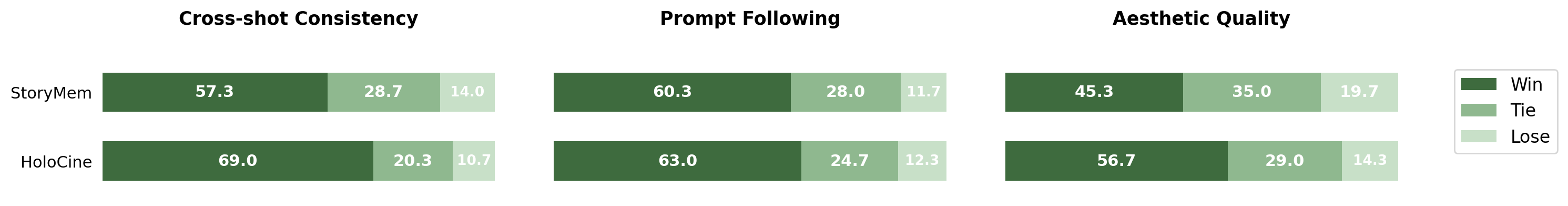} 
    \caption{User study results. We report pairwise human preference comparisons between our method and each baseline over 30 generation cases evaluated by 10 participants. For each criterion, Win, Tie, and Lose indicate the percentage of comparisons where our method is preferred over, judged comparable to, or preferred less than the corresponding baseline.}
    \label{fig:user_study}
    \vspace{-0.2cm}
\end{figure*}

To complement automatic evaluation, we conduct a user study with 10 participants on 30 diverse generation cases. For each case, evaluators perform pairwise comparisons between our method and each baseline, including StoryMem~\cite{storymem} and HoloCine~\cite{holocine}. The comparisons are conducted under three criteria: cross-shot consistency, prompt following, and aesthetic quality. For each criterion and each baseline, evaluators choose whether our method is better, comparable, or worse, corresponding to Win, Tie, and Lose, respectively. The reported percentages are computed over 300 pairwise judgments, corresponding to 30 cases evaluated by 10 participants.

As shown in Fig.~\ref{fig:user_study}, our method is consistently preferred over both baselines across all three criteria. For cross-shot consistency, our method achieves a win rate of 57.3\% against StoryMem and 69.0\% against HoloCine, indicating stronger long-term subject preservation across shots. For prompt following, our method also obtains clear advantages, with win rates of 60.3\% against StoryMem and 63.0\% against HoloCine. In terms of aesthetic quality, our method achieves win rates of 45.3\% and 56.7\% against StoryMem and HoloCine, respectively, while also receiving substantial tie rates. These results suggest that our method improves perceived cross-shot coherence and prompt adherence while maintaining competitive or better visual quality.
```

\subsection{Ablation Studies}
\label{ablation}


Tab.~\ref{tab:ablation} validates the effectiveness of the proposed reconstruction objective and disentangled memory. Adding subject reconstruction improves inter-shot subject consistency from $0.7227$ to $0.7489$, indicating that explicit reconstruction encourages the model to preserve and reuse subject identity from memory. However, after introducing disentangled memory, the inter-shot score slightly decreases to $0.7338$. This is because the memory bank is no longer optimized only for the immediate next shot; instead, it may introduce memories of other recurring subjects that are not directly related to the next shot. Such additional subject evidence can slightly weaken adjacent-shot similarity, but it provides richer identity cues for long-range consistency.

Despite the slight drop in inter-shot consistency, the disentangled memory module improves other consistency metrics, including semantic consistency, background consistency, intra-shot subject consistency, and inter-scene subject consistency. These results suggest that disentangled memory has a positive effect on global subject preservation. By separating long-term identity evidence from short-term contextual cues, the model can better maintain recurring subjects across the whole video rather than overfitting to local next-shot continuity.




\begin{table*}[t]
    \centering
    \caption{Ablation studies on the individual components of Memento. Starting from the
    baseline, we \textit{cumulatively} add each module and report aesthetic quality,
    semantic consistency (story and shot-level), background consistency, and subject
    consistency across different granularities. The best results are highlighted in
    \textbf{bold}.}
    \label{tab:ablation}
    \resizebox{\textwidth}{!}{
    \begin{tabular}{l c cc c ccc}
        \toprule
        \multirow{2}{*}{Method} & \multirow{2}{*}{Aesthetic} & \multicolumn{2}{c}{Semantic Consistency} & \multirow{2}{*}{\shortstack{Background \\ Consistency}} & \multicolumn{3}{c}{Subject Consistency} \\
        \cmidrule(lr){3-4} \cmidrule(lr){6-8}
        & & Story & Shot & & Inter-shot & Intra-shot & Inter-scene \\
        \midrule
        Baseline                                          & 0.4937 & 0.2793 & \underline{0.2681} & 0.9744 & 0.6606 &  \underline{0.8146} & 0.6692 \\
        + learnable query                                 & \underline{0.4975} &  0.2793 & 0.2622 & \underline{0.9752} & 0.7227 & 0.7616 & \underline{0.7177} \\
        + recon. task                                     & 0.4901 & \underline{0.2799} & 0.2679 & 0.9723 & \textbf{0.7489} & 0.7243 & 0.7082 \\
        + disentangled memory              & \textbf{0.4977} & \textbf{0.3063} & \textbf{0.2893} & \textbf{0.9805} & \underline{0.7338} & \textbf{0.8578} & \textbf{0.7268} \\
        \bottomrule
    \end{tabular}
    }
\end{table*}

\subsection{Advanced Capabilities}
\label{capability}

\begin{figure*}[t]
    \centering
    \includegraphics[width=0.85\textwidth]{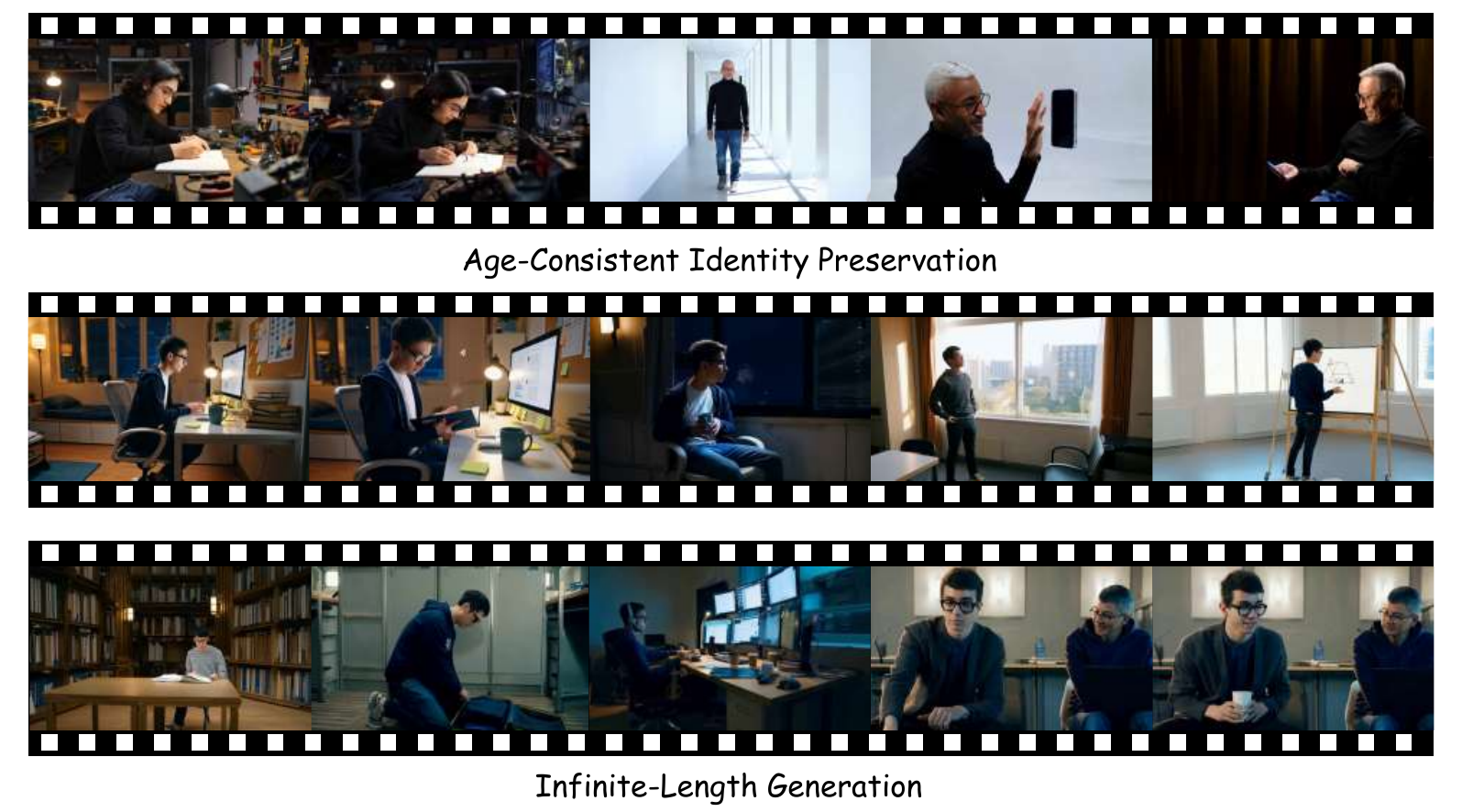} 

    \caption{Advanced capabilities of Memento.
Top: our method can generate narratives involving age variation while preserving a recognizable subject identity across different life stages. 
Bottom: our autoregressive multi-shot generation framework scales to long-form video synthesis; we further generate a 5-minute video to demonstrate its ability to maintain coherent story progression and subject consistency over extended duration.}
    \label{fig:app}
    \vspace{-0.5cm}
\end{figure*}

Beyond standard multi-shot generation, Memento exhibits two advanced capabilities, as shown in Fig.~\ref{fig:app}. First, it supports age-consistent identity preservation: the model can generate narratives where the same character changes age while maintaining stable identity cues such as facial structure, hairstyle tendency, and overall appearance. This suggests that Memento does not merely copy a fixed reference, but separates persistent identity from controllable temporal variations. Second, our autoregressive generation with dynamic memory update scales naturally to long-form synthesis. In a 5-minute generation example, Memento maintains coherent subject appearance, scene progression, and visual quality across many connected shots, demonstrating practical scalability toward minute-level narrative video generation.

\section{Conclusion}

We present Memento, a subject-reconstruction-guided framework for consistent long video generation. By combining dual-query memory retrieval with memory-based subject reconstruction, Memento explicitly preserves long-term identity evidence while maintaining short-term contextual continuity for next-shot generation. Together with a subject-aware data curation pipeline, Memento achieves strong cross-shot subject consistency, narrative coherence, and visual quality, while naturally extending to long-form generation without architectural modification. 

\subsection{Limitation and Future Work}
Memento has two limitations. First, autoregressive generation may propagate errors when degraded shots are written into memory. Second, physical plausibility remains limited by the video backbone, which lacks explicit modeling of gravity, object permanence, and rigid-body dynamics. Future work will study robust memory filtering and physics-aware generation.

\bibliography{neurips_2026}
\bibliographystyle{plain}


\newpage
\appendix
\section{Appendix}
\section{Training Details}

\section{Additional Data Details}
\label{app:data_statistics}

We provide additional statistics of the curated training data in this section. The dataset contains 2,033 video sequences and 20,227 clips in total. Each sequence contains 9.95 clips on average, with a minimum of 6 clips and a maximum of 16 clips. The average sequence duration is 45.9 seconds, the median duration is 42.9 seconds, and the duration ranges from 30 to 65 seconds.

For data annotation, we use Qwen3-VL-8B for caption generation, PaddleOCR for OCR detection, and ByteTrack for identifying reconstruction frames. Qwen3-VL-8B is called 26,252 times in total across three captioning tasks: sequence captioning, clip captioning, and reconstruction captioning. The detailed caption statistics are shown in Table~\ref{tab:caption_stats}.

\begin{table}[h]
\centering
\caption{Statistics of the curated dataset and generated captions. Caption length is measured by character count.}
\label{tab:caption_stats}
\begin{tabular}{lrrrrr}
\toprule
\textbf{Type} & \textbf{Count} & \textbf{Min} & \textbf{Max} & \textbf{Mean} & \textbf{Median} \\
\midrule
Sequence caption & 2,030 & 80 & 1,485 & 543 & 513 \\
Clip caption & 20,189 & 152 & 999 & 459 & 450 \\
Reconstruction caption & 3,992 & 33 & 641 & 130 & 117 \\
\bottomrule
\end{tabular}
\end{table}

\section{Evaluation Metrics}
\label{app:metrics}

We provide detailed definitions of the evaluation metrics used in our benchmark. Our evaluation covers semantic consistency, background consistency, subject consistency, aesthetic quality, and cross-shot/scene consistency. Unless otherwise specified, all feature vectors are L2-normalized before cosine similarity is computed.

\subsection{Semantic Consistency}

Semantic consistency measures whether the generated video is aligned with the textual prompts. We use ViCLIP~\cite{wang2022internvideo} with the OpenGVLab/ViCLIP-B-16-hf checkpoint as the video-text encoder.

\paragraph{Global semantic consistency.}
For each generated scene, we uniformly sample frames from the whole scene, with 24 frames used by default. The sampled frames are divided into clips of 8 frames. Each clip is fed into ViCLIP to extract a video feature, and the features of all clips are averaged and L2-normalized to obtain the scene-level video representation. We concatenate all video prompts within the scene into a single text description and encode it with ViCLIP to obtain the text feature. The global semantic consistency score is computed as the cosine similarity between the scene-level video feature and the text feature.

\paragraph{Shot-level semantic consistency.}
For shot-level evaluation, each shot is processed independently. We uniformly sample frames from each shot and divide them into 8-frame clips. ViCLIP is then used to extract clip-level video features, which are averaged to form the shot-level video representation. The corresponding prompt of the shot is encoded as the text feature. We compute the cosine similarity between the shot-level video and text features, and report the average score over all shots.

\subsection{Background Consistency}

Background consistency measures the temporal stability of scene appearance and background layout. We use CLIP ViT-B/32 with the openai/clip-vit-base-patch32 checkpoint as the image encoder.

Given a generated video or scene, frames are sampled with a fixed interval denoted as \texttt{CLIP\_SAMPLE\_INTERVAL}. Each sampled frame is encoded by CLIP into an image feature, followed by L2 normalization. We compute the cosine similarity between adjacent sampled frames and average the similarities as the background consistency score.

We report this metric at two levels. The scene-level background consistency, denoted as intra-scene background consistency, is computed within each scene. The video-level background consistency, denoted as intra-video background consistency, is computed over all frames concatenated across the whole story.

\subsection{Subject Consistency}

Subject consistency evaluates whether recurring subjects maintain stable visual identity over time. We use DINOv2 ViT-B/14~\cite{oquab2023dinov2} with the facebookresearch/dinov2 checkpoint as the visual feature extractor.

Frames are sampled with a fixed interval denoted as \texttt{DINO\_SAMPLE\_INTERVAL}. Each sampled frame is encoded by DINOv2, and the extracted feature is L2-normalized. We compute two complementary similarity terms. The first is the average cosine similarity between consecutive sampled frames, which measures short-term temporal smoothness. The second is the average cosine similarity between each sampled frame and the first sampled frame, which measures long-range identity preservation. The final subject consistency score is defined as:
\begin{equation}
    S_{\text{subject}} = \frac{1}{2}
    \left(
    S_{\text{consecutive}} + S_{\text{first-frame}}
    \right),
\end{equation}
where $S_{\text{consecutive}}$ denotes the mean adjacent-frame similarity and $S_{\text{first-frame}}$ denotes the mean similarity to the first frame.

Similar to background consistency, this metric can be computed at the scene level and video level, corresponding to intra-scene subject consistency and intra-video subject consistency, respectively.

\subsection{Aesthetic Quality}

Aesthetic quality evaluates the visual appeal of generated frames. We use OpenAI CLIP ViT-L/14 as the image encoder, followed by the LAION aesthetic MLP predictor. For each frame, the CLIP feature is extracted, L2-normalized, and then fed into the aesthetic MLP to predict an aesthetic score.

The original aesthetic score lies in the range $[1, 10]$. We normalize it to $[0, 1]$ as follows:
\begin{equation}
    S_{\text{aesthetic}} = \frac{S_{\text{raw}} - 1}{9}.
\end{equation}
The final aesthetic score is obtained by averaging the normalized frame-level scores over all generated frames. We also record the standard deviation to reflect the stability of aesthetic quality across frames.

\subsection{Inter-Shot Consistency}

Inter-shot consistency measures whether visual and identity information is preserved across adjacent shots. We consider three variants.

\paragraph{ViCLIP-based inter-shot consistency.}
For each shot in a scene, we uniformly sample \texttt{INTER\_SHOT\_NUM\_FRAMES} frames and extract a shot-level video feature using ViCLIP. We then compute cosine similarities between adjacent shot pairs and report the average similarity as the inter-shot consistency score.

\paragraph{DINOv2-based inter-shot subject consistency.}
We also compute an identity-focused inter-shot metric using DINOv2. For each shot, sampled frame features are extracted by DINOv2 and mean-pooled into a single shot-level representation. The cosine similarities between adjacent shot representations are then averaged.

\paragraph{Character-grouped inter-shot consistency.}
To better evaluate recurring subject consistency, we further compute a character-grouped inter-shot score. Based on the \texttt{[Person X]} tags in the prompts, shots are grouped by the characters they contain. For each character group, we compute pairwise ViCLIP cosine similarities among all shots involving the same character. The final score is obtained by averaging the pairwise similarities within each character group and then averaging across all character groups.

\subsection{Intra-Shot Consistency}

Intra-shot consistency evaluates temporal coherence within the same scene or shot. In our benchmark, it corresponds to the scene-level versions of subject consistency and background consistency, namely intra-scene subject consistency and intra-scene background consistency. These metrics measure whether subjects, backgrounds, and visual layouts remain stable across frames within a local temporal segment.

\subsection{Inter-Scene Consistency}

Inter-scene consistency evaluates whether subject and visual identity can be preserved across scene transitions. We use ViCLIP to extract one feature vector for each scene by encoding the frames of the entire scene. We then compute cosine similarities between adjacent scene pairs and report their average as the inter-scene consistency score. A higher score indicates better long-range consistency across scene boundaries.


\newpage

\end{document}

%% file: section/intro.tex
\section{Introduction}

\begin{figure*}[t]
    \centering
    \includegraphics[width=\textwidth]{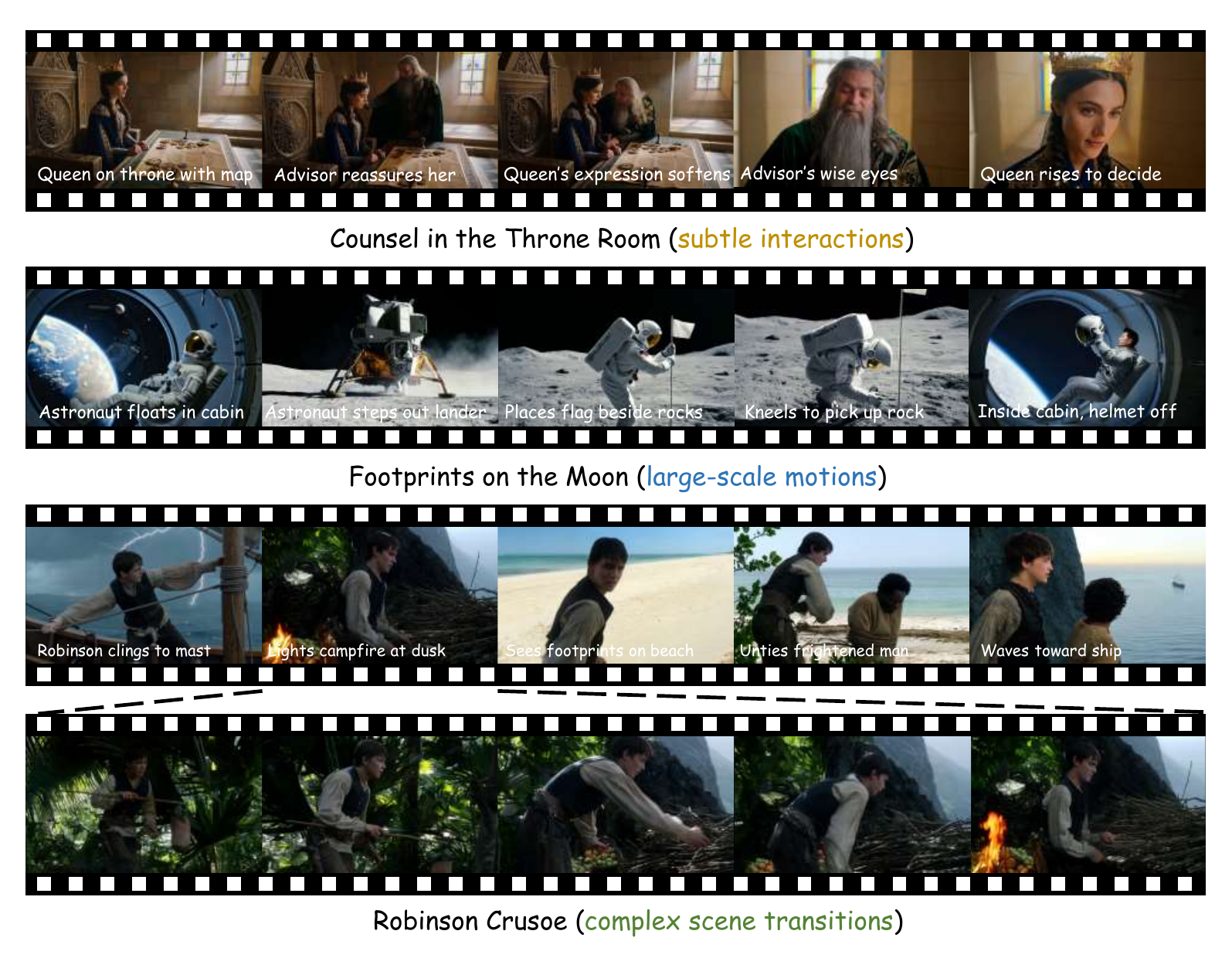} 
    \caption{
We show representative results covering progressively challenging settings. The examples illustrate Memento's ability to preserve different subject identities and background coherence over long temporal horizons. Whether handling subtle interactions, large-scale motions, or complex scene transitions in a survival story, our method ensures natural motion and consistent character appearance across multiple shots.
}

    \label{fig:teaser}
\end{figure*}

Recent diffusion-based video generation models have achieved remarkable progress in visual fidelity, motion realism, and text alignment~\cite{diffusion, dit, blattmann2023stable, rombach2022high, wan, kong2024hunyuanvideo, yang2024cogvideox, veo3, kling, sora2, yu2025context, seedance2026seedance}. Despite these advances, generating long-form and story-driven videos remains challenging.
A long video is not merely an extended sequence of visually plausible frames, it requires recurring subjects to maintain their identities across multiple shots, viewpoints, poses, motions, and scene transitions. However, subtle changes in face, clothing, body shape, or object appearance can accumulate over time, causing subject drift and breaking narrative coherence.

Existing long video generation methods typically address this challenge through temporal decomposition.
Specifically, storyboard-based approaches~\cite{storydiffusion, xie2024dreamfactory, zhao2024moviedreamer, zheng2024videogen, hu2024storyagent, he2025cut2next, xiao2025captain, zhang2025shouldershot, long2024videostudio, iclora} synthesize sparse keyframes before animating them into clips, while joint multi-shot methods~\cite{lct, holocine, wu2025cinetrans, qi2025mask, kara2025shotadapter, cai2025mixture, jia2025moga} generate multiple shots within a shared diffusion context. 
Although these strategies improve cross-shot coordination, they either leave individual clips largely independent or remain constrained by context length and escalating long-range attention costs.
Alternatively, memory-conditioned autoregressive methods~\cite{storymem, onestory} offer a scalable alternative by generating videos shot by shot with compact historical memory. 
However, their memory selection mechanisms are usually driven by generic relevance, visual saliency, or short-term context compression, rather than explicit identity preservation. As generation proceeds, identity-critical evidence of recurring subjects may be gradually diluted.

We contend that subject consistency in long video generation is fundamentally an identity grounding problem. 
Existing memory-conditioned methods typically use historical frames as references for next-shot generation, where the training signal primarily rewards plausible local continuation. 
While such supervision can improve temporal coherence in the short term, it does not explicitly guarantee that the memory retains fine-grained identity evidence for recurring subjects. 
Consequently, the model may learn to exploit memory for scene layouts or visual continuity, while failing to preserve the critical appearance cues required to recognize the same subject across distant shots.

To directly supervise identity preservation, we introduce subject reconstruction as an auxiliary training objective. Our key intuition is straightforward: \emph{if a memory bank truly preserves a subject, the model should be capable of recovering the subject from memory}. 
We therefore require the model to reconstruct target subject appearances using only historical memory and the global story caption, without relying on direct visual prompts.
This transforms subject consistency from an implicit expectation of next-shot generation into an explicitly verifiable objective, thereby forcing the memory to encode and retrieve identity-relevant cues such as face, clothing, body structure, and other stable appearance attributes.

This reconstruction-guided perspective further suggests that a single entangled memory is suboptimal for long video generation.
Subject reconstruction and next-shot generation depend on different forms of historical information. Specifically, reconstruction benefits from long-range identity evidence that should remain stable across scene changes and distant shots.
In contrast, next-shot generation relies more heavily on short-range contextual cues, such as recent layout, motion tendency, and camera continuity.
Consequently, when both objectives share the same memory selection policy, visually salient recent context can easily dominate the memory, while sparse but identity-critical evidence may be suppressed. 
This motivates a disentangled memory design that separately retrieves long-context subject evidence for identity grounding and short-context visual references for local shot generation.

Based on this design, we propose \textbf{Memento}, a subject-reconstruction-guided framework for consistent long video generation. Given a global story caption and a sequence of per-shot captions, Memento generates videos autoregressively at the shot level while maintaining a fixed-length historical memory bank. At each generation step, the memory candidate pool is formed from previous memory tokens and visual features of the most recent shot. 
A dual-query memory mechanism is then introduced to retrieve complementary historical references: story-conditioned long-context queries select subject-relevant memories for identity preservation, whereas shot-conditioned short-context queries capture local references for the upcoming shot.
The retrieved memories are used jointly for diffusion-based next-shot generation and memory-based subject reconstruction, enabling scalable long video generation without full global attention. 
To support training, we further develop a subject-aware data curation pipeline that generates global story captions, local shot captions, and reconstruction target captions with consistent, pronoun-free subject descriptions, which significantly reduces ambiguity in subject-level supervision.

Our contributions are summarized as follows:
\begin{itemize}
\item We propose a memory-based subject reconstruction objective that explicitly enforces identity preservation for recurring subjects during long video generation.

\item We introduce \textbf{Memento}, a novel subject-reconstruction-guided framework, which leverages a dual-query memory mechanism to separate the retrieval of long-context subject evidence and short-context visual cues, facilitating scalable shot-level autoregressive synthesis.

\item We construct a subject-aware cinematic data curation pipeline with consistent, pronoun-free subject descriptions and reconstruction targets.

\item Extensive experiments demonstrate that Memento achieves superior long-term subject consistency, cross-shot coherence, and visual quality across diverse long-form scenarios.

\end{itemize}